\pdfoutput=1

\documentclass[11pt]{article}

\usepackage{ACL2023}

\usepackage{times}
\usepackage{latexsym}

\usepackage[T1]{fontenc}

\usepackage[utf8]{inputenc}

\usepackage{microtype}

\usepackage{inconsolata}

\usepackage{todonotes}

\usepackage{cleveref}
\usepackage{multirow}
\usepackage{soul}
\usepackage{xspace}

\newcommand{\cd}{CO$_2$\xspace}
\newcommand{\roberta}{RoBERTa\xspace}

\newcommand{\ngrams}{$n$-grams\xspace}

\newcommand{\codelink}{\url{https://github.com/kamruzzaman15/Sentiment-Analysis-Feature-Extraction-}}

\title{Efficient Sentiment Analysis: A Resource-Aware Evaluation of Feature Extraction Techniques, Ensembling, and Deep Learning Models}

\author{Mahammed Kamruzzaman \\
  University of South Florida \\
  Tampa, FL, USA 33620 \\
  \texttt{kamruzzaman1@usf.edu} \\\And
  Gene Louis Kim \\
  University of South Florida \\
  Tampa, FL, USA 33620 \\
  \texttt{genekim@usf.edu} \\}

\begin{document}
\maketitle
\begin{abstract}

Huge amounts of sentiment-rich data are produced on social networks, the rate of which is growing year by year. As such, a global view of sentiment on social media platforms is only possible at scale through automation. The huge data sizes also mean that sentiment analysis systems must balance both accuracy and computing resource requirements. Typical NLP research maximizes accuracy, while other metrics are overlooked and thus older models are forgotten. Prior models are easily forgotten despite their possible suitability in settings where large computing resources are unavailable or relatively more costly. In this paper, we perform a broad comparative evaluation of document-level sentiment analysis models with a focus on resource costs that are important for the feasibility of model deployment and general climate consciousness. Our experiments consider different feature extraction techniques, the effect of ensembling, task-specific deep learning modeling, and domain-independent large language models~(LLMs). We find that while a fine-tuned LLM achieves the best accuracy, some alternate configurations provide huge~(up to $24,283\times$) resource savings for a marginal~(<1\%) loss in accuracy. Furthermore, we find that for smaller datasets, the differences in accuracy shrink while the difference in resource consumption grows further.\footnote{This paper was accepted and presented at the Proceedings of the 11th International Workshop on Natural Language Processing for Social Media, held in conjunction with \textbf {AACL 2023}.}

\end{abstract}


\section{Introduction}
\label{sec:introduction}

The wider NLP community in recent years has seen a trend in growing models and data sizes to improve performance. This trend is most clearly demonstrated by the progression of large language models over the past few years, from ELMo~\cite{peters-etal-2018-deep}, followed by BERT~\cite{devlin-etal-2019-BERT}, RoBERTa~\cite{liu-etal-2019-roberta}, GPT-3~\cite{brown-etal-2020-large}, and LaMDA~\cite{thoppilan-etal-2022-lamda}, just to name several notable examples, though this is by no means the only area of NLP where this is happening. This trend leaves resource efficiency and sustainability as an afterthought, a luxury that social media platforms may not have due to the sheer amount of data produced on social networks. Resource-intensive models not only cost more but also have a greater carbon impact---a consideration that is increasingly important to make in a climate-conscious society.
\begin{figure}
\centering
\includegraphics[width=1.0\linewidth]{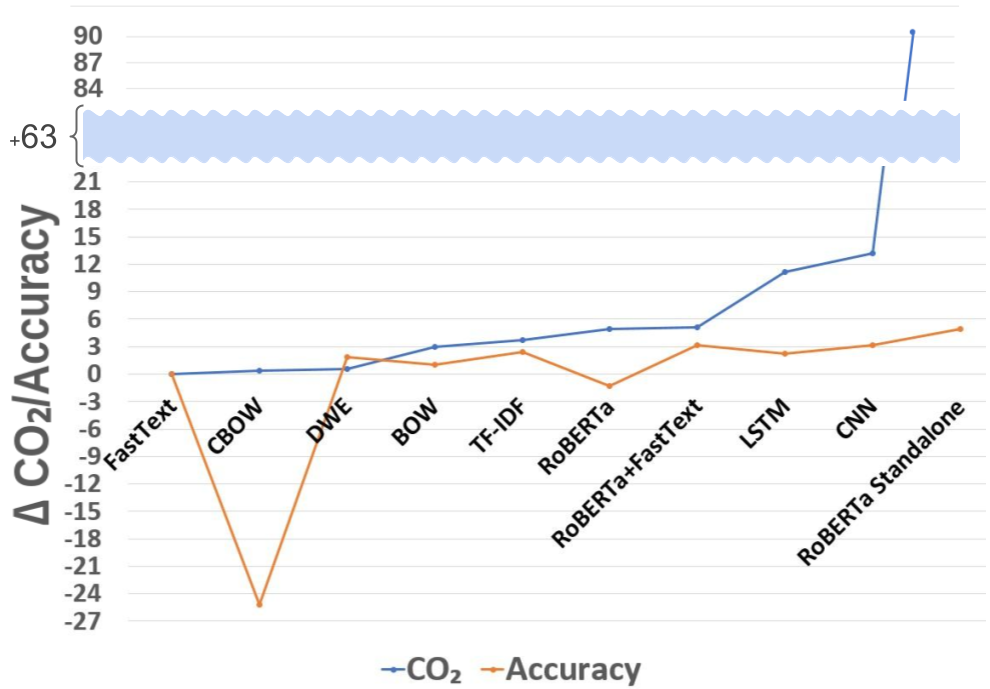}
\captionof{figure}{Change in accuracy and estimated grams of \cd emitted from various feature extraction models and the standalone fine-tuned \roberta model on the IMDB dataset \cite{IMDB}. Plotted values are averages across all considered classifiers. The changes in values~($\Delta$\cd~/~accuracy) were computed relative to FastText, which had the lowest \cd of all feature extractors we considered. For full results see \Cref{tab:IMDB-tests}.}
\label{fig:imdb-delta-plots}
\end{figure}

Ensemble methods are another common, though more old-fashioned, technique for attaining performance improvements. Ensemble learning combines the results of several models and frequently achieves state-of-the-art~(SOTA) results in benchmarks across NLP domains. These models are often only outperformed by large language models that have been pretrained on vast amounts of data. Notable examples in NLP include the Graphene algorithm for AMR Parsing~\cite{hoang-etal-2021-ensembling}, the SQuAD 2.0 question answering task~\cite{rajpurkar-etal-2018-know} where ensemble models consistently beat single-models of comparable complexity. In these performance-topping scenarios, the ensemble learning methods lead to increased computing costs and associated greenhouse gas emissions due to the multiple models that make up the ensemble.

In this paper, we investigate the relationships between system performance, computing costs, and environmental costs while considering both large pretrained models and ensembling methods to find an accurate view of the trade-offs that we make when we select one model over another. Performance analysis in the literature often limits its view to only measuring the quality of the output (e.g., accuracy), ignoring both real-world resource constraints such as time and memory requirements and the relative environmental impact of using one model over another. 

We use the task of document-level sentiment analysis as a case study. This ensures that the experiments stay within the scope of a single paper and that we avoid an explosion of experimental settings. 
Social networks produce huge amounts of sentiment-rich data in various forms, such as comments, reviews, and customer service chat, which are difficult to parse manually. Sentiment analysis is a critical tool for providing a global view of sentiment on a social media platform. 

In our investigation, we use three different review datasets and consider nine feature extraction methods, two ensembling methods, and three standalone models. The wide range of datasets and methods ensures we can identify broad trends and make replicable conclusions. 

This paper's contributions are the following.
\begin{itemize}
    \item We investigate the relationship between performance, resource use, and environmental impact in sentiment analysis by measuring accuracy, end-to-end runtime, memory usage, energy expenditure, and generated \cd estimates for a broad range of models that previous literature has shown to be appropriate for this task. 
    
    \item We find that although modern neural systems can outperform other methods in accuracy, this comes at a considerable cost in other resources (time, memory, and energy). The performance advantage of large neural models over other methods shrinks on smaller datasets, making alternatives more appealing.

    \item For most of the scenarios that we tested, we find that FastText with a Support Vector Machine (SVM) classifier or a frozen RoBERTa model (no finetuning) alongside FastText and an SVM classifier achieves strong performance while greatly reducing runtime and energy expenditure.
\end{itemize}
\Cref{fig:imdb-delta-plots} demonstrates all three points for feature extractors on the IMDB review dataset. It shows how accuracy and carbon emissions change across feature extractors, including the extreme carbon emissions cost of the fine-tuned \roberta model and the balanced accuracy and emissions of the RoBERTa+FastText feature extractor.

\section{Related Work}
\label{sec:related-work}

\citet{strubell-etal-2019-energy} revealed that recent NLP advancements have come with significant financial, energy, and environmental costs. \citet{hershcovich-etal-2022-towards} suggest that addressing this issue requires an emphasis in promoting climate transparency in research. In this vein, \citeposs{pranesh-shekhar-2020-analysis} analysis clearly reveals resource efficiency differences in several text classification models. Their analysis, however, did not include now-prevalent LLMs and only considered memory efficiency. \citet{bannour-etal-2021-evaluating} brought \cd measurement into such model comparison and recommended ML CO2~\cite{lacoste-etal-2019-quantifying} as a convenient carbon tracking tool. ML \cd has since developed into codecarbon~\cite{lottick2019nergy}, which we use for our experiments.


Previous studies in sentiment analysis have investigated a wide range of methodologies, including traditional machine learning models, deep machine learning models, 
and various feature extraction strategies.
Dataset-based feature extraction techniques such as bag-of-words~(BOW), TF-IDF, and \ngrams had historically been popular in sentiment analysis. Similarly, FastText, GloVe, and word2vec were popular pretrained word embeddings in sentiment analysis and other NLP classifications alike. 
\citet{zhou2022text} proposed a model that uses the double word embedding~(DWE) method for sentiment classification, where GloVe and word2vec are used in combination for input. 
\citet{alharbi2021evaluation} and \citet{goularas-etal-2019-evaluation} focused on pretrained word embedding models with neural networks. \citet{alharbi2021evaluation} compared GloVe, word2vec, and FastText word embedding models with various RNN architectures.
\citet{goularas-etal-2019-evaluation} considered GloVe and word2vec embeddings with multiple CNN and RNN architectures. They proposed a method of combining multiple CNNs with a biLSTM using GloVe embeddings as input, which they showed to outperform the other configurations in terms of accuracy.

\begin{figure*}
\centering
\includegraphics[width=0.9\linewidth]{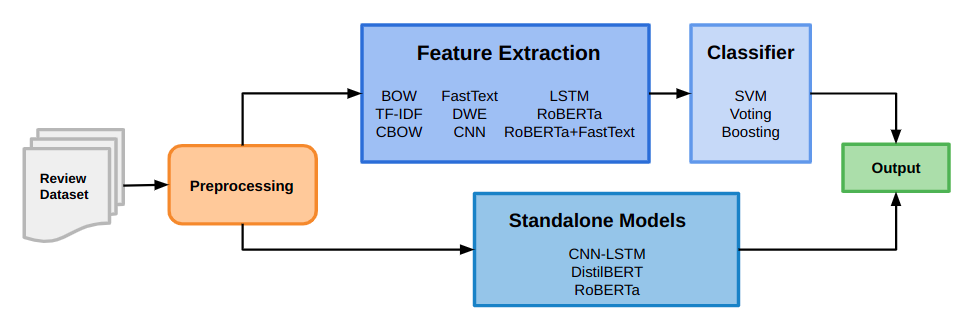}
\captionof{figure}{High-level data flow from the review dataset to the output for models composed of feature extractors with classifiers and standalone models.}
\label{fig:meth}
\end{figure*}

\section{Methodology}
\label{sec:methodology}

We separate our models into two broad categories. One category contains models based on an arbitrary combination of a feature extractor and a classifier. The other category contains standalone models where at least one of these two components is inherent to the model design. The input for all models is preprocessed in the same manner and all models are trained to produce a binary sentiment polarity classification. \Cref{fig:meth} illustrates the high-level data flow from the review dataset to the output for the two types of models. 
For data preprocessing details see \Cref{app:data-preprocessing}. See \Cref{app:abandoned-methods} for a discussion of feature extractors, classifiers, and standalone models that were abandoned from our final evaluation.
The remainder of this section provides brief introductions to each of the feature extractors, classifiers, and standalone models that we consider in our experiments.\footnote{The source code for our evaluations is available at \codelink.}



\subsection{Feature Extraction Methods}
\label{sec:FEM}

\paragraph{Bag of Words~(BOW).}

BOW is a classic, simple approach to converting text data into a vector that can be input into an ML model. Each document is represented by a $V$ dimensional vector, where $V$ is the vocabulary size. Each dimension value corresponds to the number of times the word appears in the text.


\paragraph{Term Frequency-Inverse Document Frequency (TF-IDF).}

TF-IDF combines the term frequency and inverse document frequency to produce a value representing the relevance of a word to a document~\cite{SALTON1988513}.
The term frequency is the number of times the word appears in the text. The inverse document frequency is the inverse of the number of documents that the word appears in. This ensures that words that are common across texts, such as function words, are not over-valued. TF-IDF unfortunately may become computationally expensive for large vocabularies.



\paragraph{Continuous Bag of Words (CBOW).}

The CBOW model encodes the meaning of words using the several surrounding context words as feedback \cite{mikolov-etal-2013-efficient}. From this, it predicts contextually correct sentences. As CBOW models are able to make use of more data during the input than the Skip-gram model~\cite{mikolov-etal-2013-efficient}, they can capture the word vectors efficiently and hence perform better than Skip-gram. We trained our CBOW model using the word2vec module in the Gensim library~\cite{rehurek_lrec}. 


\paragraph{FastText.}
\label{par:fasttext}

FastText~\cite{bojanowski2016enriching,joulin-etal-2017-bag} is a highly engineered library by Facebook for representation learning and classification exploiting subword information and quantization-based memory optimizations. We use a supervised learning algorithm to train FastText on our specific training data, which allows us to learn word embeddings from each dataset. Further details of this system are beyond the scope of this paper.


\paragraph{Double Word Embedding (DWE).}
\label{par:dwe}



DWEs were introduced by \citet{zhou2022text} as a method to obtain combine the benefits of two different word embedding schemes using a convolution layer with piece-wise max pooling. We use DWEs combining FastText and GloVe whereas \citet{zhou2022text} used GloVe and word2vec based on preliminary experiments on all combinations of the three word embedding models. Word-level GloVe vectors are mean-pooled to obtain document-level vectors. FastText vectors are obtained according to the same approach as described in the FastText section.

\paragraph{CNN.}

Our convolutional neural network~(CNN) feature extractor has the following architecture, which is described below. Each convolution layer is followed by ReLU activation, max pooling, and dropout. For training, after the final convolution layer (+ aforementioned post-convolution steps) the data is fed through two dense layers and a sigmoid. When used as a feature extractor, the feature embeddings are pulled from the last convolution layer, before the max pooling stage, but after ReLU activation. These are fed into the classifiers. For the IMDB dataset, we use two convolution layers. For the smaller restaurant and product review datasets, we only use a single convolution layer.


\paragraph{LSTM.}

Long short-term memory networks~(LSTMs) are designed to learn long-term dependencies in the input, such as those seen in text data. Our LSTM design is similar to our CNN feature extractor. Each LSTM layer is followed by a dropout layer. During training, prediction is done after two additional dense layers and a sigmoid activation. When used as a feature extractor, features are pulled from the last LSTM layer before dropout and fed into classifiers. We use two LSTM layers for the IMDB dataset and one layer for the two smaller datasets. 


\paragraph{RoBERTa.}
\label{ssec:roberta}

We use the 12th hidden layer for the [CLS] token of the RoBERTa-base model as the feature vector. When used as a feature extractor, RoBERTa is fully frozen and never trained.

\paragraph{RoBERTa + FastText.}


FastText and RoBERTa feature vectors---as described in FastText and RoBERTa section above---are concatenated to create a joint representation of the input text data.

\subsection{Classification Models}

\paragraph{SVM.}

Support vector machines~(SVMs) select the hyperplane that divides the data points into labeled classes while maintaining a margin between the hyperplane and the nearest data point(s)~\cite{708428}. For binary classification, this hyperplane is positioned to accomplish the maximum separation between the two classes, resulting in the largest possible margin. SVMs are popular classifiers, especially in machine learning application domains, due to their robustness to small dataset sizes.


\paragraph{Voting Ensemble.}

Voting ensembles take the predictions~(``votes'') from member classifiers and select the winning vote. We use a soft-voting method described by \citet{10.5555/1972514}, where each member classifier produces a probability distribution over the possible classes, and the class with the highest average probability wins. We use SVM, Random Forest~(RF), and Multinomial Na{\"i}ve Bayes~(MNB) as the member classifiers in our experiments. 


\paragraph{Boosting Ensemble.}

Boosting is a sequential ensemble process for changing
the weight of observations based on the previous classifications iteratively. When an observation is wrongly labeled, the weight of the observation rises. This eliminates dataset bias error and generates robust statistical models. During preparation, the Boosting algorithm assigns weights to each of the resulting models. For boosting, we use XGBoost (eXtreme Gradient Boosting)~\cite{10.1145/2939672.2939785}.

\subsection{Standalone Models} 

In initial experiments, we found that changing the classification head led to a consistent decrease in performance for certain neural models. 
This suggests that the feature extraction stages of these models are interwoven with the classification method over which they are trained.
In these cases, we keep both the feature extraction and classification components together without swapping out classification heads. 

\paragraph{CNN-LSTM.}

Following \citet{goularas-etal-2019-evaluation}, we use a combination of CNN and LSTM to extract the local and global characteristics in the input text. Each CNN layer is followed by ReLU activation, max pooling, and dropout. The first CNN layer takes randomly initialized vocabulary vectors as input. The LSTM layers then run on top of this. Each LSTM layer is followed by dropout. Finally, two dense layers and a sigmoid activation are used for prediction. For the IMDB dataset, we use two CNN and LSTM layers each. For the restaurant and product review datasets, we only use a single layer each.


\begin{table*} [!thbp]
\begin{center}
{\small
\begin{tabular}{ |l|l|r|r|r|r|r|  }
\hline
\multicolumn{7}{ |c| }{IMDB Review Dataset} \\

\hline
Feature Extraction Method & Classifier & Accuracy & Time (min) & Memory (MB)& Energy (Wh) & CO$_2$(g.)\\ \hline 

\multirow{3}{*}{BOW} & SVM & 87.86 & 18.75 & 178.72 & 3.66 & 1.701\\
 & Voting & 88.06 & 82.01 & 181.61 & 16.02 & 7.442\\
 & Boosting & 85.40 & 0.16 & 152.82 & 0.03 & 0.015 \\ \hline
 \multirow{3}{*}{TF-IDF} & SVM & 90.11 & 22.87 & 179.36 & 4.46 & 2.075 \\
 & Voting & 89.82 & 103.93 & 182.28 & 20.30 & 9.432 \\
 & Boosting & 85.56 & 0.22 & 154.21 & 0.04 & 0.020 \\ \hline
\multirow{3}{*}{CBOW} & SVM & 61.43 & 3.60 & 106.41 & 0.70 & 0.327 \\
 & Voting & 61.26 & 10.29 & 106.36 & 2.011 & 0.934 \\
 & Boosting & 60.10 & 2.12 & 79.07 & 0.41 & 0.192\\ \hline
\multirow{3}{*}{FastText} & SVM & 88.64 & 0.34 & 65.58 & 0.06 & 0.031 \\
 & Voting & 81.25 & 2.38 & 66.06 & 0.46 & 0.216 \\
 & Boosting & 88.40 & \textbf {0.08} & \textbf {60.69} & \textbf {0.01} & \textbf {0.007} \\ \hline
\multirow{3}{*}{DWE} & SVM & 88.72 & 5.24 & 745.46 & 1.02 & 0.476 \\
 & Voting & 86.47 & 11.22 & 732.16 & 2.19 & 1.018 \\
 & Boosting & 88.73 & 3.25 & 710.59 & 0.63 & 0.295 \\ \hline
\multirow{3}{*}{CNN} & SVM & 89.59 & 89.82 & 6519.82 & 17.54 & 8.151\\
 & Voting & 89.65 & 337.81 & 6519.87 & 65.98 & 30.655 \\
 & Boosting & 88.63 & 11.16 & 5150.24 & 2.81 & 1.013 \\ \hline
\multirow{3}{*}{LSTM} & SVM & 88.62 & 123.10 & 361.67 & 24.04 & 11.172 \\
 & Voting & 88.37 & 126.95 & 362.14 & 24.80 & 11.521 \\
 & Boosting & 87.87 & 122.45 & 355.43 & 23.92 & 11.112\\ \hline
\multirow{3}{*}{RoBERTa} & SVM & 83.88 & 54.59 & 774.55 & 10.66 & 4.953 \\
 & Voting & 85.78 & 68.07 & 732.08 & 13.29 & 6.176 \\
 & Boosting & 84.60 & 43.42 & 632.45 & 8.48 & 3.941 \\ \hline
 \multirow{3}{*}{RoBERTa + FastText} & SVM & 89.51 & 57.47 & 789.90 & 11.22 & 5.214 \\
 & Voting & 89.27 & 69.01 & 747.43 & 13.45 & 6.250\\
 & Boosting & 88.94 & 44.80 & 691.29 & 8.75 & 4.065 \\ \hline 
\hline

 \multicolumn{2}{| c |}{\textit{Standalone RoBERTa Baseline}} & \textbf {91.02} & 997.88 & 606.14 & 201.05 & 90.554 \\ \hline
\end{tabular}
}
\end{center}
\caption{\label{tab:IMDB-tests}Results of IMDB review dataset on various combinations of feature extraction methods and classifiers.}
\end{table*}

\paragraph{RoBERTa-base.}

When used as a standalone model, we jointly fine-tune all layers of the RoBERTa base model and train the classifier on each dataset. We obtain a RoBERTa base model classifier using the Huggingface~\cite{wolf-etal-2020-transformers} AutoModel feature for sequence classification.

\paragraph{DistilBERT-base-uncased.}

We use DistilBERT base uncased model as a standalone model in our experiment. This model is fine-tuned with an added classification layer in the same manner as the RoBERTa base model.

\begin{figure*}
\centering
\includegraphics[width=0.9\linewidth]{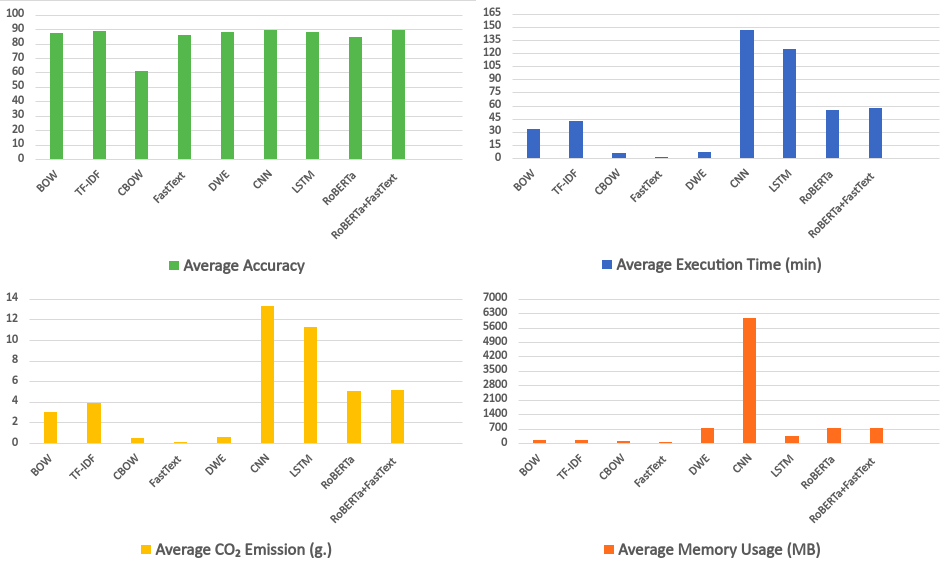}
\captionof{figure}{Feature extraction results for the IMDB dataset, averaged across classifiers.}
\label{fig:IMDB}
\end{figure*}

\section{Experimental Setup}

All experiments were conducted on a single workstation in Florida, USA.
For more details of the workstation configuration see \Cref{app:config}.

\paragraph{Datasets.}

We evaluate our models on three different binary sentiment classification datasets. 
We evaluate our model configurations on the IMDB movie review dataset \cite{IMDB}, the ``Restaurant Reviews in Dhaka, Bangladesh'' dataset~\cite{restaurant}, and the ``Grammar and Online Product Review'' dataset~\cite{product}.
Following \citet{DBLP:ZhangZL15} and \citet{10.5555/2021109.2021114}, we construct a binary classification task from restaurant and product review datasets by labeling 1-star samples as negative polarity and 5-star samples as positive polarity examples. For more details of datasets see \Cref{app:datasets}. In all of our experiments, 80\% of the data is used for training and 20\% for testing.

\paragraph{Model Parameters.}

The context window size of the CBOW model is set to 15 and the min-count is set to 1 so that it considers all words. For SVM, MNB, and RF, we use the default parameters provided by the \texttt{scikit-learn} library~\cite{scikit-learn}. For the XGBoost classifier, we use the default parameters of the \texttt{xgboost} library~\cite{10.1145/2939672.2939785}.
We adjust the non-pretrained neural models based on different sizes of the datasets. For more details about these parameters see \Cref{app:model-parameters}.

\paragraph{Training Parameters.} For the CNN-LSTM models, we choose the number of epochs between 2 and 5 depending on the dataset size. For the restaurant and product review dataset, the batch size is 32 and for IMDB, the batch size is 16. Adaptive Moment Estimation is used as an optimizer with binary cross-entropy loss. The CNN and LSTM feature extractors are only trained for 1 epoch each. The fine-tuned RoBERTa-base model is trained with a learning rate of 1e-1, accumulation steps of 2, and batch size of 8. We train it for 3 epochs using the AdamW optimizer. DistilBERT is trained with a batch size of 16. Otherwise, the parameters are the same as RoBERTa.


\paragraph{Evaluation Metric.}
To evaluate our models we measure accuracy, end-to-end execution time or total time, and memory measurement that includes the memory usage of both the model itself (model size) and the memory consumed during the forward pass or inference, electricity consumption, and \cd emissions. We use codecarbon~\cite{lottick2019nergy} to estimate the carbon emissions.

\section{Results \& Analysis}
The full results on the IMDB, restaurant, and product review datasets for various feature extraction-classifier combinations are provided in \Cref{tab:IMDB-tests}, \Cref{tab:restaurant-tests}, and \Cref{tab:product-tests}, respectively, alongside the standalone RoBERTa model. In every table, \cd emissions are measured in grams, energy consumption in Watt-hours (Wh), time in minutes, and memory in megabytes~(MB). 
\Cref{fig:IMDB} plots the average accuracy, \cd, time, and memory usage for each feature extraction technique on the IMDB dataset.
The same plot for the restaurant review dataset~(\Cref{fig:res}) is provided in \Cref{app:restaurant-figure}.
Looking at only the accuracy numbers, we find that the fine-tuned \roberta model outperforms every other configuration in every dataset. However, upon closer inspection, we find that the story is not so simple. 

\paragraph{Feature Extractors and Classifiers.} In \Cref{tab:IMDB-tests}, we see that the standalone RoBERTa model achieves the best accuracy of 91.02\%, followed by TF-IDF, FastText, and RoBERTa + FastText feature extractors with the SVM classifier, which achieve 90.11\%, 88.64\%, and 89.51\%, respectively. If we take a look at the carbon emissions, these models do considerably better than the standalone RoBERTa model. TF-IDF with SVM results in 43 times less carbon emissions with only a 0.91\% accuracy reduction. Similarly, RoBERTa + FastText with SVM produces 17 times less carbon emissions, while DWE with boosting produces 306 times less carbon emission, resulting in 1.51\% and 2.29\% accuracy reductions, respectively. When taking these other resources into consideration, FastText may be a better option than the fine-tuned \roberta model. The CNN feature extractor achieved competitive accuracy but has considerable memory requirements. It is also worth mentioning that end-to-end runtime and CO$_2$ for CNN as a feature extractor with SVM and Voting is more than the RoBERTa as a feature extractor because the feature layer of CNN is computationally intensive for its number of filters and kernels. 

\begin{table*} [!htbp]
\begin{center}
{\small
\begin{tabular}{ |l|l|r|r|r|r|r|  }
\hline
\multicolumn{7}{ |c| }{Restaurant Review Dataset} \\
\hline
Feature Extraction Method & Classifier & Accuracy & Time (min) & Memory (MB) & Energy (Wh) & CO$_2$ (g.) \\ \hline
\multirow{3}{*}{BOW} & SVM & 94.51 & 0.041 & 6.74 & 0.008 & 0.0037\\
 & Voting & 95.76 & 0.233 & 7.24 & 0.046 & 0.0211\\
 & Boosting & 94.80 & 0.008 & \textbf {6.10} & 0.002 & 0.0007 \\ \hline
 \multirow{3}{*}{TF-IDF} & SVM & 95.62 & 0.045 & 6.90 & 0.009 & 0.0040 \\
 & Voting & 95.81 & 0.248 & 7.39 & 0.048 & 0.0225 \\
 & Boosting & 94.56 & 0.008 & 6.29 & 0.002 & 0.0007 \\ \hline
\multirow{3}{*}{CBOW} & SVM & 77.88 & 0.078 & 12.69 & 0.015 & 0.0071 \\
 & Voting & 78.75 & 0.347 & 12.92 & 0.068 & 0.0315 \\
 & Boosting & 80.33 & 0.052 & 10.04 & 0.012 & 0.0054\\ \hline
\multirow{3}{*}{FastText} & SVM & 96.01 & \textbf {0.007} & 6.41 & \textbf {0.001} & \textbf {0.0006} \\
 & Voting & 92.11 & 0.101 & 6.77 & 0.020 & 0.0092 \\
 & Boosting & 95.91 & 0.011 & 6.18 & 0.002 & 0.0010 \\ \hline
\multirow{3}{*}{DWE} & SVM &  96.10 & 2.055 & 553.84 & 0.402 & 0.1866 \\
 & Voting & 93.79 & 2.259 & 552.60 & 0.442 & 0.2051 \\
 & Boosting & 95.86 & 2.008 & 550.33 & 0.393 & 0.1824 \\ \hline
\multirow{3}{*}{CNN} & SVM & 96.05 & 4.029 & 3044.51 & 0.787 & 0.3655\\
 & Voting & 95.62 & 16.168 & 3044.75 & 3.157 & 1.4666 \\
 & Boosting & 95.28 & 2.751 & 2641.68 & 0.537 & 0.2496 \\ \hline
\multirow{3}{*}{LSTM} & SVM & 94.85 & 5.652 & 62.40 & 1.104 & 0.5129 \\
 & Voting & 95.14 & 6.161 & 62.63 & 1.204 & 0.5591 \\
 & Boosting & 95.24 & 5.608 & 62.06 & 1.096 & 0.5089\\ \hline
\multirow{3}{*}{RoBERTa} & SVM & 80.24 & 5.133 & 149.62 & 1.003 & 0.4657 \\
 & Voting & 92.12 & 5.826 & 142.68 & 1.138 & 0.5286 \\
 & Boosting & 93.84 & 4.902 & 130.15 & 0.957 & 0.4447 \\ \hline
 \multirow{3}{*}{RoBERTa + FastText} & SVM & 96.53 & 5.921 & 151.01 & 1.156 & 0.5372 \\
 & Voting & 96.25 & 6.129 & 150.92 & 1.197 & 0.5561 \\
 & Boosting & 96.15 & 5.739 & 142.47 & 1.121 & 0.5207 \\ \hline
\hline

 \multicolumn{2}{| c |}{\textit{Standalone RoBERTa Baseline}} & \textbf {96.69} & 160.612 & 77.94 & 31.362 & 14.5701  \\ \hline
\end{tabular}
}
\end{center}
\caption{\label{tab:restaurant-tests} Results of restaurant review dataset on different feature extraction methods.}
\end{table*}

\begin{table*} [!htbp]
\begin{center}
{\small
\begin{tabular}{ |l|l|r|r|r|r|r|  }
\hline
\multicolumn{7}{ |c| }{Product Review Dataset} \\
\hline
Feature Extraction Method & Classifier & Accuracy & Time (min) & Memory (MB) & Energy (Wh) & CO$_2$ (g.) \\ \hline
\multirow{3}{*}{BOW} & SVM & 91.29 & 0.078 & 9.75 & 0.015 & 0.0071\\
 & Voting & 91.29 & 0.392 & 10.27 & 0.077 & 0.0356\\
 & Boosting & 91.05 & 0.012 & 8.59 & \textbf {0.002} & 0.0011 \\ \hline
 \multirow{3}{*}{TF-IDF} & SVM & 92.83 & 0.081 & 9.90 & 0.016 & 0.0074 \\
 & Voting & 92.08 & 0.403 & 10.41 & 0.079 & 0.0365 \\
 & Boosting & 91.57 & 0.012 & 8.79 & 0.003 & 0.0011 \\ \hline
\multirow{3}{*}{CBOW} & SVM & 75.88 & 0.120 & 14.83 & 0.023 & 0.0109 \\
 & Voting & 78.79 & 0.432 & 14.89 & 0.085 & 0.0392 \\
 & Boosting & 80.33 & 0.094 & 10.78 & 0.018 & 0.0085\\ \hline
\multirow{3}{*}{FastText} & SVM & 93.44 & \textbf {0.010} & 7.33 & \textbf {0.002} & \textbf {0.0010} \\
 & Voting & 90.73 & 0.136 & 7.63 & 0.027 & 0.0123 \\
 & Boosting & 93.39 & 0.014 & \textbf {6.82} & 0.003 & 0.0013 \\ \hline
\multirow{3}{*}{DWE} & SVM & 92.93 & 2.092 & 557.20 & 0.409 & 0.1899 \\
 & Voting & 90.91 & 2.346 & 554.76 & 0.459 & 0.2130 \\
 & Boosting &  93.82 & 2.070 & 551.72 & 0.405 & 0.1880 \\ \hline
\multirow{3}{*}{CNN} & SVM & 92.69 & 5.717 & 3297.52 & 1.117 & 0.5186\\
 & Voting & 92.88 & 22.072 & 3326.16 & 4.324 & 1.9965 \\
 & Boosting & 92.04 & 3.648 & 2714.40 & 0.714 & 0.3306 \\ \hline
\multirow{3}{*}{LSTM} & SVM & 92.41 & 6.006 & 65.20 & 1.173 & 0.5452 \\
 & Voting & 92.97 & 6.144 & 65.45 & 1.200 & 0.5576 \\
 & Boosting & 92.88 & 5.933 & 64.67 & 1.159 & 0.5384\\ \hline
\multirow{3}{*}{RoBERTa} & SVM & 84.64 & 5.352 & 166.12 & 1.045 & 0.4856 \\
 & Voting & 87.68 & 6.611 & 151.27 & 1.291 & 0.5998 \\
 & Boosting & 90.26 & 5.114 & 133.97 & 0.999 & 0.4640 \\ \hline
 \multirow{3}{*}{RoBERTa + FastText} 
 & SVM      & 94.14       & 6.269     & 156.73    & 1.224     & 0.5688 \\
 & Voting   & 93.49                & 6.837     & 156.28    & 1.335     & 0.6203 \\
 & Boosting & 93.53                & 6.163     & 146.63    & 1.221     & 0.5682 \\ \hline
 \hline

  \multicolumn{2}{| c |}{\textit{Standalone RoBERTa Baseline}} & \textbf {94.66} & 213.961 & 95.89 & 41.779 & 19.4092  \\ \hline
\end{tabular}
}
\end{center}
\caption{\label{tab:product-tests}Results of product review dataset on different feature extraction methods.} 
\end{table*}

\begin{table*}[!thbp]
\begin{center}
{\small
\begin{tabular}{ |l|l|r|r|r|r|r|  }
\hline
\multicolumn{7}{ |c| }{Standalone Models} \\
\hline
Dataset & Models & Accuracy & Time (min) & Memory (MB) & Energy (Wh) & CO$_2$ (g.) \\ \hline
\multirow{3}{*}{IMDB} & CNN-LSTM & 88.41 & \textbf{39.44}  & \textbf{328.97} & \textbf{7.705} & \textbf{3.579}\\
 & DistilBERT &  89.92 & 498.29 & 614.11 & 97.310 & 45.207 \\
 & RoBERTa & \textbf {91.02} & 997.88 & 606.14 & 201.054 & 90.554 \\ \hline
\multirow{3}{*}{Restaurant} & CNN-LSTM & 96.03 & \textbf{2.27} & \textbf{58.33} & \textbf{0.444} & \textbf{0.206}\\

 & DistilBERT &  96.57 & 76.01 & 71.18 & 14.845 & 6.896 \\
 & RoBERTa & \textbf {96.69} & 160.61 & 77.94 & 31.362 & 14.570 \\ \hline
\multirow{3}{*}{Product} & CNN-LSTM & 92.13 & \textbf{2.78} & \textbf{60.66} & \textbf{0.543} & \textbf{0.252} \\
 & DistilBERT &  93.89 & 105.59 & 91.53 & 20.621 & 9.580 \\
 & RoBERTa & \textbf {94.66} & 213.96 & 95.89 & 41.779 & 19.409 \\ \hline
\end{tabular}
}
\end{center}
\caption{\label{tab:standalone-tests}Results of standalone models on the three datasets.}
\end{table*}

\Cref{tab:restaurant-tests} shows similar results for the restaurant review dataset. RoBERTa achieves the highest accuracy, while FastText, DWE, and RoBERTa + FastText feature extractors with the SVM classifier achieve nearly the same performance with major carbon emission reductions. DWE with SVM produces 78 times less carbon with only a 0.59\% reduction in accuracy when compared to fine-tuned \roberta. Similarly, RoBERTa + FastText with SVM and FastText with SVM produce 27 times and 24,283 times less carbon emissions, while only losing 0.16\% and 0.68\% in accuracy, respectively. 



The results from the product review dataset~(\Cref{tab:product-tests}) further solidify our observations so far. DWE with boosting, RoBERTa + FastText with SVM, and FastText with SVM result in 103 times, 34 times, and 19,409 times less carbon emissions with only 0.84\%, 0.52\%, and 1.22\% accuracy reduction, respectively, when compared to fine-tuned \roberta.


\paragraph{Standalone Models.} The results of our standalone models are presented in \Cref{tab:standalone-tests}. For IMDB review, RoBERTa requires almost 17 hours to complete the task, consuming 201.05 Wh of electricity which produces 90.554g of CO$_2$. DistilBERT takes almost 9 hours and emits 45.207 g CO$_2$ which is comparatively much lower than the RoBERTa model.

\paragraph{CNN vs. CNN-LSTM.} We also find that using CNN as a feature extractor produces slightly better results than the CNN-LSTM model, though at the cost of higher memory use.\footnote{We also tried CNN and LSTM as standalone models but the results were worse than the CNN-LSTM model.} For example, in the IMDB review dataset, CNN with the voting ensemble attains 89.65\% accuracy whereas the CNN-LSTM standalone model achieves 88.41\% accuracy.




\section{Conclusion \& Future Work}
We contextualized the accuracy of sentiment analysis systems within their computing resource requirements: runtime, maximum memory use, energy expenditure, and estimated \cd emissions. As expected, a finetuned LLM achieves the highest accuracy in all three datasets. However, this comes at the cost of tens to thousands of times the cost in terms of other resources (time, memory, energy, and \cd) when compared to other configurations which result in accuracy reductions of less than 1\%. We find that the FastText feature extractor with an SVM classifier, in particular, achieves good accuracy scores while having minimal resource requirements.

As such, for the vast majority of use cases, we recommend this configuration (concatenating the frozen RoBERTa embeddings to FastText (RoBERTa+FastText) w/ SVM) 
where one must balance resource costs with output quality. 
It provides strong accuracy scores without incurring the extreme time and energy costs of fine-tuning RoBERTa. 
Although our experiments only considered the task of sentiment analysis, we expect that the major patterns would carry over to other classification tasks to the degree that they have similar dataset features (e.g., text length, dataset size, difficulty, etc.). 
We encourage others to test this hypothesis and provide insights into which techniques best balance output quality with resource usage in other NLP domains.


Due to our own resource constraints, our experiments do not include the most cutting-edge LLM models (e.g., GPT-3.5, LLaMa, etc.). The use and deployment of these models are qualitatively different (e.g., prompting, RLHF, etc.) from those we considered and thus are outside of the scope of this paper. We leave this extension of our experiments for future work.


\section*{Acknowledgements}
This project was fully supported by the University of South Florida.

\bibliography{anthology,custom}
\bibliographystyle{acl_natbib}

\appendix

\section{Data Preprocessing}
\label{app:data-preprocessing}

For data preprocessing, we normalize all the texts by converting them into lowercase, removing URLs, HTML tags, email addresses, non-alphabetic characters, all the special characters, and stop words, tokenizing the text data, and lemmatizing the text. We also transform contractions to their expanded full form to maintain uniformity.

\section{Abandoned Methods}
\label{app:abandoned-methods}

Early in our experiments, we also considered i)~GloVe, ii)~Skip-gram, and iii)~lower layers (non-12th) of RoBERTa-base as feature extractors, bagging as a classifier, and ALBERT as a standalone model.  However, quite quickly we found that these methods consistently gave poor accuracy results or results that were not qualitatively different from a feature extractor already considered in our experiments.

\section{Workstation configuration}
\label{app:config}

GPU: RTX 3090 (24GB VRAM); RAM: 32 GB (2x 16GB Corsair CMW32GX4M2Z3600C18); CPU: AMD Ryzen 9 5900X 12-Core Processor. 

\section{Dataset Details}
\label{app:datasets}
The IMDB movie review is a balanced dataset of 25k samples for each positive and negative label. The restaurant review dataset is collected from 338 restaurants in Dhaka, Bangladesh and the product review dataset is procured from 1,000 different products. These original datasets use 5-star rating systems and we use 1-star samples as negative polarity and all 5-star samples as positive polarity examples. This leads to 8,621 positive reviews and 2,241 negative reviews in the restaurant review dataset. As the product review dataset is highly imbalanced and we seek to investigate the effect of dataset size in our experiments, we further subsample the positive reviews in the product review dataset down to 6,981 positive reviews, while keeping all 3,701 negative reviews.

\section{Model Parameters Details}
\label{app:model-parameters}
Regarding the non-pretrained neural models, we use a dropout rate of 0.2, maximum sequence lengths of 512, and LSTM layers with 128 units.
For the restaurant and product review datasets, the vocabulary size of embedding layers is 10,000 and the CNN layers have a kernel size of 3, with 128 filters.
For the IMDB review dataset, the kernel size and number of filters in the first convolution layer are 5 and 300, respectively, while those for the second convolution layer are 3 and 100. The size of the vocabulary is 30,000. 

Additionally, we also try different parameters of SVM, XGBoost, MNB, and RF but the default one got the more prominent result compared to others. For example, when we use RF in our voting ensemble method, we also try the number of estimators=200, random state=1200, and criterion='entropy'. 

The RoBERTa base model comes pretrained with 12 transformer layers, 12 attention heads, and a hidden layer size of 768. DistilBERT base uncased model has 6 layers with 12 attention heads. Other tested hyperparameters for LLMs are represented in \Cref{tab:hyperparameter-tests}

\begin{table*}
\centering
\begin{tabular}{|c|c|c|}
\hline
Hyperparameter & Tested Values & Optimal Value \\
\hline
Optimizer & Adam, Nadam, SGD & Adam \\
\hline
Epochs & 2, 3, 4 & 3  \\
\hline
Learning Rate & 1e-1, 5e-1, 2e-1 & 1e-1 \\
\hline
Batch Size & 4, 8, 16 & 8 \\
\hline
\end{tabular}
\caption{Tested Hyperparameters for LLMs other than Specified Particularly}
\label{tab:hyperparameter-tests}
\end{table*}

\section{Average Result Diagram for Restaurant Review Dataset}
\label{app:restaurant-figure}
\begin{figure*}
\centering
\includegraphics[width=0.9\linewidth]{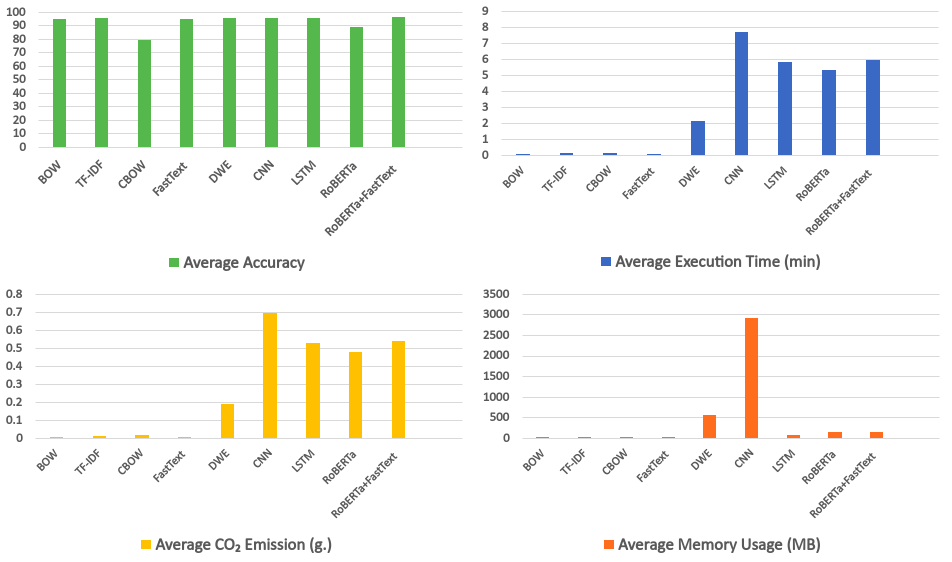}
\captionof{figure}{Average Results Diagram of  Restaurant Review Dataset}
\label{fig:res}
\end{figure*}


\end{document}